\documentclass[lettersize,journal]{IEEEtran}
\usepackage{amsmath,amsfonts}
\usepackage{algorithm}
\usepackage{array}
\usepackage[caption=false,font=normalsize,labelfont=sf,textfont=sf]{subfig}
\usepackage{textcomp}
\usepackage{stfloats}
\usepackage{url}
\usepackage{verbatim}
\usepackage{graphicx}
\usepackage{cite}
\usepackage{color}
\usepackage{xcolor}
\usepackage{algorithm}
\usepackage{algpseudocode}

\def\NoNumber#1{{\def\alglinenumber##1{}\State #1}\addtocounter{ALG@line}{-1}}
\newcommand{\rightcomment}[1]{\hfill\textcolor{gray}{\% #1}}
\renewcommand{\comment}[1]{\NoNumber{\textcolor{gray}{\% #1}}}

\begin{document}

\title{Accelerating Reinforcement Learning for Autonomous Driving using Task-Agnostic and Ego-Centric Motion Skills}

\author{Tong Zhou$^{1,2,*}$, Letian Wang$^{3,*}$, Ruobing Chen$^{1}$, Wenshuo Wang$^{4}$, Yu Liu$^{1,\dagger}$

\thanks{* denotes equal contribution, the listing order is random.}
\thanks{$\dagger$ denotes corresponding author}
\thanks{$^{1}$SenseTime Research, Beijing, China.}%
\thanks{$^{2}$Department of Electronic Engineering, The Chinese University of Hong Kong, Shatin, N.T., Hong Kong SAR, China.}%
\thanks{$^{3}$University of Toronto, Toronto, ON, CA (lt.wang@mail.utoronto.ca).}%
\thanks{$^{4}$California PATH, UC Berkeley, CA, USA.}%
}

\maketitle

\begin{abstract}
Efficient and effective exploration in continuous space is a central problem in applying reinforcement learning (RL) to autonomous driving. Skills learned from expert demonstrations or designed for specific tasks can benefit the exploration, but they are usually costly-collected, unbalanced/sub-optimal, or failing to transfer to diverse tasks. However, human drivers can adapt to varied driving tasks without demonstrations by taking efficient and structural explorations in the \textit{entire} skill space rather than a limited space with task-specific skills. Inspired by the above fact, we propose an RL algorithm exploring \textit{all} feasible motion skills instead of a limited set of task-specific and object-centric skills. Without demonstrations, our method can still perform well in diverse tasks. First, we build a task-agnostic and ego-centric (TaEc) motion skill library in a pure motion perspective, which is diverse enough to be reusable in different complex tasks. The motion skills are then encoded into a low-dimension latent skill space, in which RL can do exploration efficiently. Validations in various challenging driving scenarios demonstrate that our proposed method, TaEc-RL, outperforms its counterparts significantly in learning efficiency and task performance. 
\end{abstract}

\section{Introduction}
\IEEEPARstart{R}{einforcement} learning (RL) has achieved great success in various domains \cite{silver2016mastering,vinyals2019grandmaster,liu2021motor} by learning via trial-and-error during constant interaction with the environment. However, the learning efficiency will quickly decay in large-dimension environments such as autonomous driving in urban traffic due to un-informed exploration in continuous action space and the sparse and delayed rewards. As a result, existing plain RL algorithms are often data inefficient to necessitate a large amount of experience, and failing in complex tasks or environments.

Toward such learning efficiency and performance concerns, one great attempt is to accelerate RL algorithms via expert demonstrations. Many existing works contribute this direction, such as pre-training policy via imitation learning  \cite{peters2008reinforcement}\cite{pan2020navigation}, regularizing and augmenting the reward during RL training \cite{nair2018overcoming,zhu2018reinforcement,wu2021shaping}, and injecting demonstrations into the replay buffer \cite{hester2018deep}\cite{vecerik2017leveraging}. Ideally, these methods can overcome exploration challenges and even further surpass the performance of expert policies. However, such expert demonstrations are usually (i) expensive and labor-intense to collect and annotate if unavailable, (ii) unbalanced in distribution and hardly optimal-guaranteed, and (iii) struggling to transfer to new tasks as the demonstrations are environment-conditioned or task-specific.

\begin{figure}[t]
    \centering
    \includegraphics[width=0.49\textwidth]{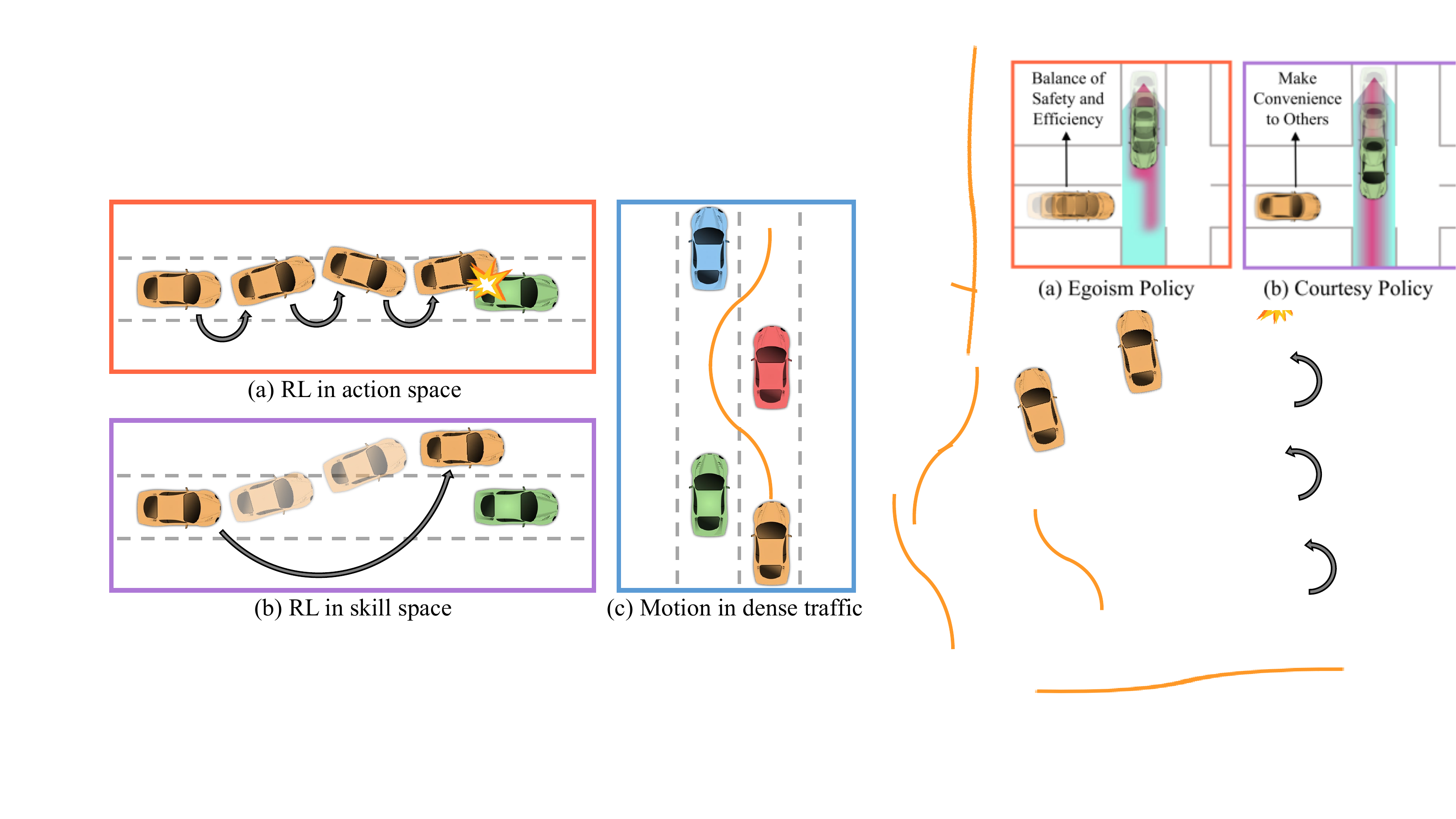}
    \caption{RL agents (e.g. autonomous vehicles) taking exploration (a) over the raw action space will results in inconsistent action sequences, or (b) over the skill space can generate a sequence of consistent low-level actions. (c) In dense traffic, autonomous vehicles need to consider the relationships of surrounding vehicles to generate desired motions, which is too complicated to manually design from a task or object view.}
    \label{fig:motivation}
\end{figure}

Behavioral science reveals that human behavior is in nature temporally extended \cite{jing2004construction}, 
whose low-level actions should be regarded as results of \textit{skill executions} rather than a decision space to explore. As in Fig~\ref{fig:motivation}(b), human drivers' sequence of consistent low-level actions are instinctive muscle responses to a decided high-level intention like overtaking the front car. The temporally-extended skills enables efficient learning with structured exploration and accelerated reward encountering. In comparison, most existing RL algorithms explore in raw actions and result in inconsistent action sequences, which rarely reflect driving intentions. For instance, autonomous vehicles might move in a wiggling way when purely making explorations in the action space, thereby failing to achieve specific tasks such as overtaking a leading car, as shown in Fig.~\ref{fig:motivation}(a). As a result, a number of works explicitly exploited skills to encourage structural exploration and accelerate reward encountering. Some works learn or distinguish skills in offline motion dataset \cite{pertsch2020accelerating}\cite{lynch2020learning}. Nevertheless, it is still hard to guarantee that all essential skills are learned or covered in the dataset. Other works manually design delicate skills such as task-specific decision hierarchies \cite{wang2022transferable,merel2017learning,wang2021hierarchical} or object-centric parameterized primitives \cite{deo2018would,dalal2021accelerating,wang2021learning}. These methods succeed in certain settings of well-defined tasks such as overtaking one specific car. However, these task-specific and object-centric skill design limits the flexibility and expressiveness of motions, which is particularly indispensable in complex environments. Fig.~\ref{fig:motivation}(c) illustrates a multi-agent scenario in which an autonomous car needs to navigate in a dense traffic flow. In such multi-agent settings, the relationships of surrounding vehicles need to get integrated into motion generation, which is usually too complicated to design manually.
To tackle the above-mentioned limitations, we proposed TaEc-RL (RL with Task-agnostic and Ego-centric motion skills), an RL algorithm exploring all feasible \textit{motion} skills. Our method can perform well in diverse environments and tasks without demonstrations or delicate task designs.
First, we design a task-agnostic and ego-centric (TaEc) motion skill library capable of covering all possible dynamic-feasible ego motions. 
With sufficient expressiveness and flexibility, the skills only need to be defined once and then can be reused in diverse tasks. Also, note that the design of motion skills is effortless as it has been well investigated in sampling-based motion planning communities \cite{li2020IFAC,gu2017improved,wang2021socially}. Then, we distill these motion skills into a low-dimensional latent skill space, in which the autonomous vehicle can be trained with RL efficiently and effectively. By learning over all possible motion skills, our method retains the potential to solve diverse, complex tasks like driving in multi-agent settings.

In summary, our contributions are threefold: 
\begin{itemize}
    \item Designing a task-agnostic and ego-centric motion skill library, which is general-purpose to cover diverse motion skills and can be reused across tasks.
    \item Distilling motion skills into a latent skill space and modifying the RL algorithm to explore in the latent skill space to encourage efficient and effective learning.
    \item Demonstrating that our method achieves efficient and effective learning for autonomous driving in three challenging dense-traffic scenarios.
\end{itemize}
 
\section{Related Work}
\subsection{State Space Motion Primitive}
 Given boundary state constrains (such as final position, orientation and velocity), motion primitive methods \cite{mueller2015computationally,howard2007optimal,gu2017improved} can find valid trajectories quickly.
\cite{mueller2015computationally} proposed a closed-form solution of motion primitives for quadrocopters, so it didn't take non-holonomic properties into consideration.
\cite{howard2007optimal} used Newton's method to optimize parameterized control signals to generate dynamic-feasible trajectories for vehicles. Moreover, by building a lookup table storing amount of trajectories, the users can directly pick the most suitable one as initial guess or final output.
\cite{li2020IFAC, gu2017improved} used path-velocity decomposition method to generate spatial curve and speed profile independently and combine them together.

\subsection{Composite Task Learning in Reinforcement Learning}
Performing composite and long-term tasks is very essential ability in complex task in reinforcement learning area. \cite{lee2018composing} predefined a discrete primitive libraries,
and at each decision time, the policy chooses a primitive index to execute the corresponding task sequence. However, each task in the discrete primitive libraries needs to be specified or collected by intense labor. \cite{pertsch2020accelerating,kipf2019compile} extract tasks using unstructured data without human annotation. \cite{kipf2019compile} explored the embedding of action sequence into a discrete or continuous latent space. \cite{pertsch2020accelerating} learns a prior of action sequence based on current environment states to help exploration in the reinforcement learning task. These two methods rely heavily on the quality of expert data and will fail in those driving scenarios lacking efficient expert data.

\subsection{Reinforcement Learning in Autonomous Driving}
Reinforcement learning is widely used for autonomous driving in some complex scenarios. \cite{saxena2020driving} proposed a model free reinforcement learning method to negotiate with other agents in dense traffic. \cite{brito2021learning} proposed an interactive planner providing a reference velocity, and a MPC module that outputs optimal sequence of control commands minimizing a cost function. \cite{cao2020reinforcement} manually collected expert data of different driving styles to train an hierarchical framework, to solve near accident problems.

\begin{figure}[t]
    \centering
    \includegraphics[width=0.49\textwidth]{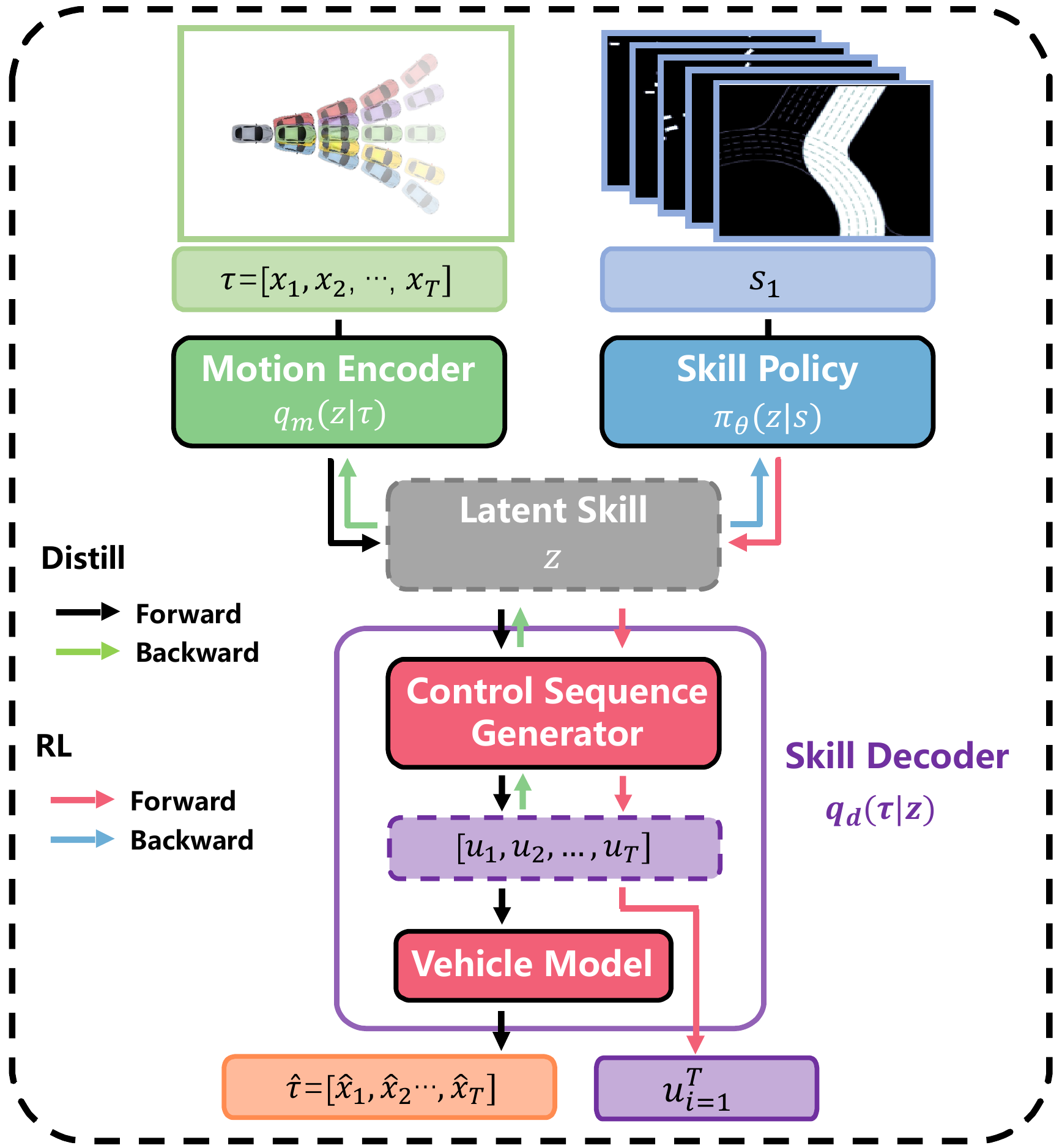}
    \caption{Illustration of the proposed pipeline. In skill distilling procedure, the TaEc motion skills are distilled into a latent skills space by a reconstruction process with the motion encoder $q_m(z|\tau)$ and the skill decoder $q_d(\tau|z)$. One sample drawn from the latent skill space represents a abstract motion, which can be recovered to a motion skill by the skill decoder $q_d(\tau|z)$.  In RL procedure,
     the skill policy $\pi_\theta(z|s)$ is trained over the latent skill space $z$ by keeping the skill decoder $q_d(\tau|z)$ fixed. The decoder outputs a series of low-level actions $\mathbf{u}^T_{i=1}$ (control signals), which will be executed sequentially at next T steps.}
    \label{fig:pipeline}
\end{figure}

\begin{figure}[htbp]
    \centering
    \includegraphics[width=0.48\textwidth]{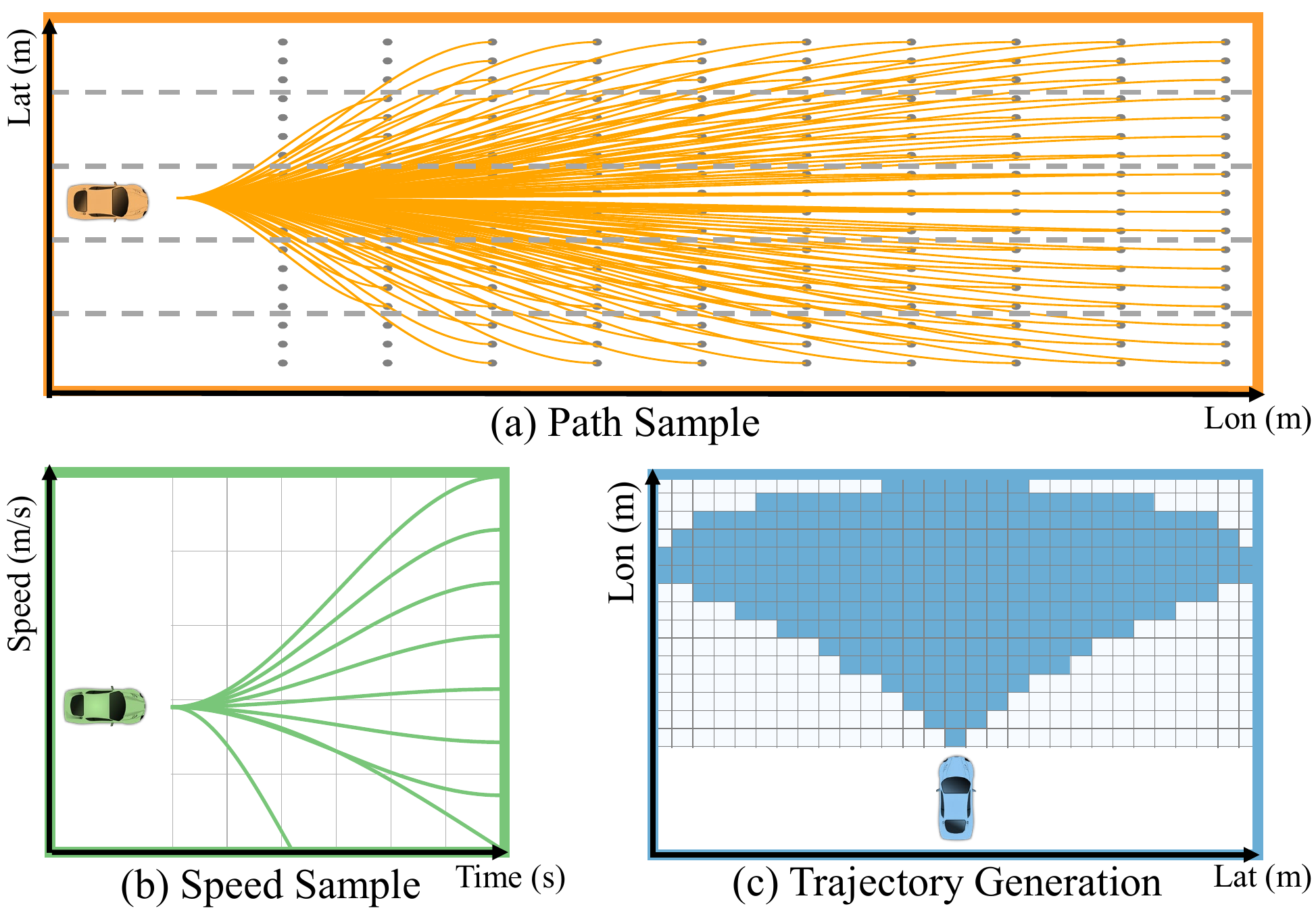}
    \caption{The TaEc motion skill library generation procedure. (a) Nodes sampling along roads to generate diverse paths. (b) Terminal speed sampling in a fixed horizon to generate diverse speed profiles. (c) Grid distribution of endpoints of all TaEc motion skill trajectories.}
        \label{fig:primitive}
\end{figure}

\section{Approach}
We aim to overcome the challenges of efficient and effective RL in continuous action space, which is a Markov Decision Process defined by a tuple $\{\mathcal{S}, \mathcal{A}, \mathcal{T}, \mathcal{R}, \gamma\}$ of states, actions, transition probabilities, rewards, and discount factor. The pipeline of our proposed TaEc-RL is illustrated in Fig.~\ref{fig:pipeline}.
First, a task-agnostic and ego-centric motion skill library $\mathcal{L}$ is designed to cover diverse motion skills, each motion skill  $\mathbf{\tau}=\{\mathbf{x_t}\}^T_{t=0}$ with a fixed horizon $T$. The motion skill is then distilled as a latent skill $z$ in a latent space $\mathcal{Z}$ by a reconstruction process with a \textbf{motion encoder} $q_m(z|\mathbf{\tau})$ and a \textbf{skill decoder} $q_d(\mathbf{\tau}|z)$. This allows conducting efficient and effective RL in the temporally-extended skill space
with a \textbf{skill policy} $\pi_\theta(z|s)$. The proposed method will be detailed in the following sections.

\subsection{Ego-Centric Task-Agnostic Motion Skill Library}
\label{sec:primitve}
We first define a task-agnostic and ego-centric (TaEc) motion skill library from a pure ego-motion view.
Specifically, we take a sampling-based motion planning approach to generate flexible and diverse motion skills. The TaEc motion skill library generation procedure consists of four steps: 1) path generation - spatial sampling method to specify the shape of the curve; 2) speed profile generation - temporal sampling method to specify the variation of speed given a time horizon; 3) raw trajectory generation - compose paths and velocity profiles together; 4) post-process the raw trajectory to get standard-form skill trajectory.
To conveniently catch up, we briefly revisit the procedures as follows.

\subsubsection{Path Generation}
The path is generated by connecting start nodes and terminal nodes in the road. Each node contains three dimensions of longitude, latitude and heading angle.
The start node is the origin of ego-vehicle's coordinate system, and the end nodes are uniformly sampled along longitude, latitude and heading angle dimensions.
For each start-end node pair, we use the state-lattice method \cite{gu2017improved} to generate a path.
The diversity of generated paths reproduces spatial motion such as lane-following, lane-change and turn around.
Fig.~\ref{fig:primitive}(a) is an example of sampling procedure along longitude and latitude (the heading angle is fixed to 0 in this figure) of terminal nodes.

\subsubsection{Speed Profile Generation} 
The speed profile can be represented as a third-order polynomial with respect to time.
\begin{equation}
    v(t) = q_0 + q_1 * t + q_2 * t^2 + q_3 * t^3,
\end{equation}
The polynomial are calculated by the initial speed, acceleration, time horizon and final speed.
The variety of speed changes can cover temporal intentions such as accelerating, decelerating and emergent stop.
As in Fig.~\ref{fig:primitive}(b), we sample the final speeds under a specific horizon given a case that initial speed is fixed.

\subsubsection{Raw Trajectory Generation}
The generated paths and velocity profiles are composed into raw trajectories as \cite{gu2017improved} to achieve spatial-temporal planning.
Each raw trajectory can be represented by a series of vehicle states, $[\mathbf{x_1}, \mathbf{x_2}, ..., \mathbf{x_{T_h}}]$, and each vehicle state is a tuple $\mathbf{x_t} = \{x_t, y_t, \phi_t, v_t\} {\forall} t \in \{1, ..., T_h\}$, where $x_t$ and $y_t$ are the longitude and latitude position, $\phi_t$ the heading angle and $v_t$ the speed with respect to the ego vehicle's current coordinates.

The raw trajectory is still gapped from composing the motion skill. One reason is the horizon $T_h$ is so long that the trajectory may contain multiple skills. The other reason is the unbalance of generated trajectory, for example, straight-line drivings occupy a substantial proportion of the trajectory set. As a result, the raw trajectories need to be processed to get TaEc Motion Skill.

\subsubsection{TaEc Motion Skill Generation} 
The Post-processing consists of two operations: Slicing and Filtering.

\begin{enumerate}
\item \textbf{Slicing.}
We adopt a time-based sliding window mechanism to slice the long-horizon trajectories into skills.
The long-horizon trajectory with length $T_h$ is mapped into sliding windows with length $T$ and sliding interval $T/2$. The divided piece with horizon $T$ is denoted as a skill trajectory, which contains different skill information at different time steps.

\item \textbf{Filtering.} 
We then perform a filtering operation by building a lookup table.
Generally, the state of the terminal node and the arc length of the trajectory is considered the most representative feature of a skill trajectory. So we discretize the 5-dimensional keys forming a look-up table. All the skill trajectories with similar features will be hashed in the same key. Lastly, each key in the table only keeps one trajectory that best matches all its contents, and the redundant ones will be filtered out.
\end{enumerate}
Finally, the TaEc skill library is approached and can be represented as $\mathbf{\tau} = [\mathbf{x_1}, \mathbf{x_2}, ..., \mathbf{x_T}]$.
 The grid occupancy map of endpoints $\mathbf{x_T}$ in TaEc motion skills is shown in Fig.~\ref{fig:primitive}(c).

\subsection{Latent Skill Space Distilling}
\label{sec:skill distill}

The TaEc skill library is then distilled into a low-dimensional latent skill space to generalize deployments of RL-based autonomous vehicles in complex and varied tasks. As shown in Fig~\ref{fig:pipeline}, a generative model consisting of a motion encoder $q_m(z|\mathbf{\tau})$ and a skill decoder $q_d(\mathbf{\tau}|z)$, is proposed. The former compresses trajectory information into low dimension latent variables, and the latter restores the trajectory from the variables. Here, to make sure the decoded trajectory is kinematically solvable for a vehicle, we propose to use vehicle kinematic model to convert the output of skill decoder into reasonable and reachable trajectories.

\subsubsection{Skill Embedding}
The TaEc motion skill library is supposed to be embed into latent skill space for the RL policy to use, under the effect of motion encoder $q_m(z|\mathbf{\tau})$ and skill decoder $q_d(\mathbf{\tau}|z)$. The motion encoder $q_m(z|\mathbf{\tau})$ takes motion skill $\mathbf{\tau}$ as input and outputs the parameters for Gaussian distribution of the latent skill $z$. One sample from the latent skill distribution represents one abstract behavior. The skill decoder $q_d(\mathbf{\tau}|z)$ reconstructs the motion skill $\hat{\mathbf{\tau}}$ from the sampling results of the latent skill distribution. The training of this model utilizes the following evidence lower bound (ELBO):
\begin{equation}\label{eq:ELBO}
    \mathbb{E}_{q_m} \bigg[\underbrace{\log q_d(\tau \vert z)}_{\text{reconstruction}} - \beta \big(\underbrace{\log q_m(z \vert \tau) - \log p(z)}_{\text{regularization}}\big) \bigg],
\end{equation}
with two types of losses in terms of reconstruction and regularization. The reconstruction loss is designed to rebuild the motion skill $\mathbf{\tau}$ from the latent skill $z$, and the regularization loss is designed to compact the latent space so as to make exploration efficiently. The prior $p(z)$ is set to be a unit Gaussian~$\mathcal{N}(0, I)$, and the $\beta$ balances the two loss terms.

To make the reconstruction temporally available, we use an LSTM model to encode and decode the motion skill. The encoder interactively inputs vehicle states yielding an embedding at the end. The skill decoder outputs a sequence of control actions at every time, which are denoted as $[\mathbf{u_1,u_2,...u_T}]$, given initial state $\mathbf{x_0}$. They are then used to generate trajectories satisfying kinematic constrains by the vehicle model.

\subsubsection{Vehicle Model} 
In the skill decoder, we use the bicycle model as the vehicle kinematic model, described as follows:
\begin{equation}\label{eq:Bicycle}
\begin{aligned}
    &\dot{x} = v \cdot cos(\phi + \beta) \\
    &\dot{y} = v \cdot sin(\phi + \beta) \\
    &\dot{\phi} = \frac{v}{l_r} \cdot sin(\beta) \\
    &\dot{v} = u^a \\
    &\beta = \arctan(\frac{l_r}{l_f+l_r}tan(u_\delta))
\end{aligned}
\end{equation}
where $\beta$ is the velocity angle, $l_r$ and $l_f$ are the distances of the rear and front tires from the gravity center of the vehicle. The state of the vehicle can be represented as $\mathbf{x} = \{x, y, \phi, v\}$, 
and the vehicle control input $\mathbf{u}$ is composed of the forward acceleration $u^a$ and steering angle $u^\delta$, $\mathbf{u} = [u^a, u^\delta]$. The vehicle model is used to convert the given control inputs into trajectory states in skill decoder.
The constrains such as the highest speed, the acceleration limit and the range of steering angle is considered in the vehicle model, in order to guarantee the trajectory is dynamic feasible.

For each time step $t \in \{1,2,...,T\}$, the vehicle model will propagates one action $\mathbf{u_t}$ at state $\mathbf{x_{t-1}}$ into vehicle model in equation \ref{eq:Bicycle} to generate a new state $\mathbf{x_{t}}$. The procedure will repeat $T$ times to generate a whole trajectory $\hat{\mathbf{\tau}} = [\hat{\mathbf{x_1}}, \hat{\mathbf{x_2}}, ..., \hat{\mathbf{x_T}}]$, which is the final output of the skill decoder $q_d(\mathbf{\tau}|z)$. The reconstruction loss consists of three parts, including the position error, the velocity error and angle error between the motion skill trajectory $\tau $ and generated trajectory $\hat{\tau}$ ,at each time step $t \in \{1,2,...,T\}$.

In light of this reconstruction, the latent skill space can represent diverse and flexible task-agnostic and ego-centric motion skills. The skill decoder $p_d(\mathbf{\tau}|z)$ will then be fixed and reused to generate future behaviors from a sample of the latent skill space. Moreover, the decoder can also be reused in vehicle with different dynamics as the vehicle model's constrains are easy to consider.

\subsection{RL with Exploration in Skill Space}
Within the latent skill space $\mathcal{Z}$, the RL agents can conduct structured and temporally-extended exploration to accelerate reward encountering. Fig.~\ref{fig:pipeline} illustrates the general idea of our proposed method. Specifically, instead of directly learning a policy over raw actions $\pi(a|s)$ (here raw action $a$ is a single control signal $\mathbf{u_1}$), we learn a policy that outputs latent skill variables $\pi_{\theta}(z|s)$ which is then decoded to motion skill by the fixed skill decoder $q_d(\mathbf{\tau}|z)$. Each motion skill is tracked for a fixed length of $T$ steps before next skill is sampled. This process follows a typical semi-MDP process with temporal abstraction and succeeds in learning for long-horizon tasks \cite{merel2018neural,merel2020catch,sutton1999between,bacon2017option}.  The horizon $T$ in RL is consistent with the skill learning of Section~\ref{sec:skill distill} and the motion skill library generation of Section~\ref{sec:primitve}. The perfect execution length might depend on tasks, environments, and circumstances, and many works attempt to learn policies with a flexible length \cite{pertsch2020keyframing,kipf2018compositional,shankar2019discovering}. In this paper, we empirically found that policies with a fixed length can reach a satisfying performance; hence, the learning for replanning triggering strategy will be as the future work. 

\begin{figure}[t!]
    \centering
    \includegraphics[width=0.45\textwidth]{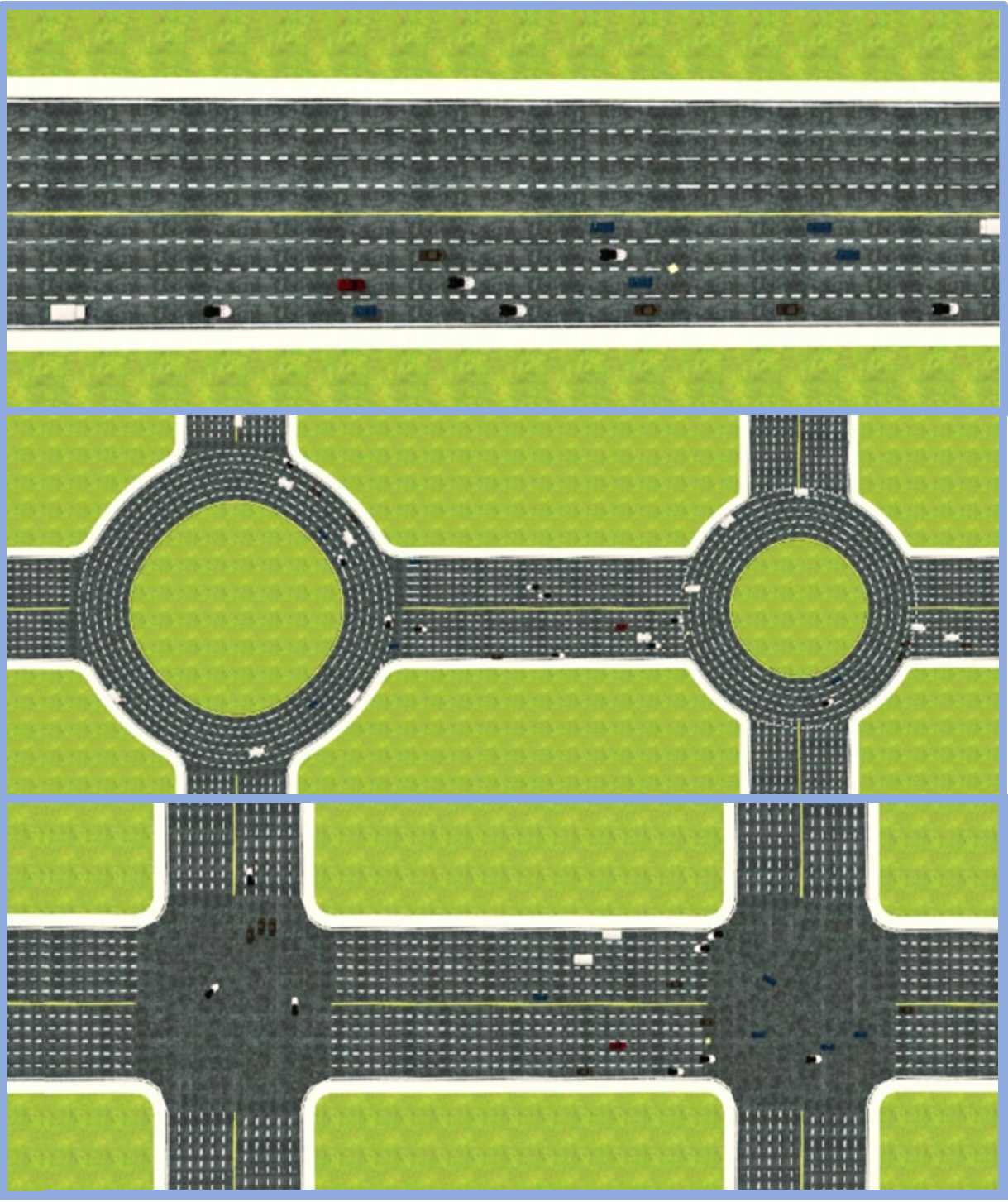}
    \caption{ Three scenarios of our experiments listed from top to down: highway, roundabout and intersection. The ego vehicle needs to reach the assigned destination with less time but no collisions or task failures such as driving out of road. }
    \label{fig: case study}
\end{figure}

To encourage exploration and enhance robustness to disturbances, we take a maximum-entropy RL \cite{ziebart2010modeling,levine2018reinforcement} to train the skill policy $\pi_{\theta}(z \vert s)$ to maximize the objective:
\begin{equation}
    J = \mathbb{E}_\pi \bigg[\sum_{i=1}^N \gamma^i \tilde{r}(s_i,z_i) + \alpha \mathcal{H}\big(\pi_{\theta}(z \vert s)\big)\bigg],
\end{equation}
where $\sum_{i=1}^N \gamma^i \tilde{r}(s_i, z_i)$ is the discounted reward returned from the environment after tracking the motion skill for $N$ decision steps ahead. This term aims to encourage the autonomous vehicle to reach destination with shortest time while penalizing jerks, collisions, and driving out of road. The entropy term $\mathcal{H}\big(\pi_\theta(z \vert s)\big)$ is designed to encourage exploration by maximizing the negated KL divergence between the latent skill space distribution (i.e., outputs of the skill policy) and a uniform distribution $U(z)$:
\begin{equation}
\mathcal{H}\big(\pi_{\theta}(z \vert s)\big) = -\mathbb{E}\big[\log \pi_{\theta}(z|s)\big] \propto -D_{\text{KL}}\big(\pi_{\theta}(z|s), U(z)\big).
\end{equation}
Specifically, we modified Soft Actor-Critic algorithm \cite{haarnoja2018soft}\cite{haarnoja2018sacapplication} to implement our idea. The entire learning procedures of TaEc RL is shown in Algorithm \ref{alg:taec_rl}.

\begin{algorithm}[t]
\begin{algorithmic}[1]
\caption{ TaEc RL} %
\label{alg:taec_rl}
\State \textbf{Input:} Motion skill library $\mathcal{L}$, discount $\gamma$, target divergence $\delta$, learning rates $\lambda_{\pi}, \lambda_Q, \lambda_\alpha$, target update rate $m$.
\State Initialize motion encoder $q_m(z \vert \tau)$, skill decoder $q_d(\tau \vert z)$, skill policy $\pi_\theta(z_t\vert s_t)$, critic $Q_\phi(s_t, z_t)$, target network $Q_{\bar{\phi}}(s_t, z_t)$, replay buffer $\mathcal{D}$

\For{each iteration}\rightcomment{Latent Skill Space Distilling}
\State Sample a skill trajectory $\tau$ from $\mathcal{L}$
\State $z \sim q_m(z \vert \tau)$; $(\{\mathbf{u_i}\}^T_{i=1}, \hat{\tau}) \sim q_d(\tau \vert z)$
\State Update $q_m, q_d$ according to Equation (\ref{eq:ELBO})
\EndFor

\For{each iteration}\rightcomment{RL in Latent Skill Space}
\For{every $T$ environment step}
\State $z_t \sim \pi_\theta(z_t \vert s_t)$ \rightcomment{sample skill latent variable}
\State ($\{\mathbf{u_i}\}^T_{i=1}, \hat{\tau}) \sim q_d(\tau \vert z_t)$ \rightcomment{generate skill}
\State $s_{t^\prime} \sim p(s_{t+T},r_{t+T} \vert s_t, \{\mathbf{u_i}\}^T_{i=1})$ \rightcomment{state transition}
\State $\tilde{r}(s_t, z_t) = \sum_{i=1}^{T}r_{t+i}$ \rightcomment{reward calculation}
\State $\mathcal{D} \leftarrow \mathcal{D} \cup \{s_t, z_t, \tilde{r}(s_t, z_t), s_{t^\prime}\}$ \rightcomment{replay buffer}
\EndFor

\For{each gradient step}
\State $\bar{Q} = \tilde{r}(s_t, z_t) + \gamma \big[ Q_{\bar{\phi}}(s_{t^\prime}, \pi_\theta(z_{t^\prime} \vert s_{t^\prime})) + \alpha \mathcal{H}\big(p_\theta(z_{t^\prime} \vert s_{t^\prime})\big)]$ \rightcomment{compute Q-target}
\comment{update policy network parameter}
\State $\theta \leftarrow \theta - \lambda_\pi \nabla_\theta [ Q_{{\phi}}(s_{t}, \pi_\theta(z_{t} \vert s_{t})) + \alpha \mathcal{H}\big(p_\theta(z_{t} \vert s_{t})\big)  ] $ 
\comment{update critic network parameter}
\State $\phi \leftarrow \phi - \lambda_Q \nabla_\phi \big[ \frac{1}{2}\big(Q_\phi(s_t, z_t) - \bar{Q} \big)^2 \big]$
\comment{update alpha}
\State $\alpha \leftarrow \alpha - \lambda_\alpha \nabla_\alpha \big[ \alpha \cdot ((\mathcal{H}\big(p_\theta(z_{t} \vert s_{t}) - \delta) \big]$ 
\comment{update target network parameter}
\State $\bar{\phi} \leftarrow \tau \phi + (1 - m) \bar{\phi}$ 
\EndFor
\EndFor
\State \textbf{return} trained policy $\pi_\theta(z_t \vert s_t)$ and skill decoder $q_d(\tau \vert z)$
\end{algorithmic}
\end{algorithm}

\section{Experiment}
In this section, we will answer the following
questions to evaluate our proposed TaEc-RL performance: (i) Can the distilled latent skill space represent diverse motion skills? (ii) Can the exploration in the skill space accelerate RL learning? (iii) Can our method transfer across different environments?

\subsection{Experiment Setting}
\subsubsection{Environment and task} 
\begin{figure}[htb]
\centering
\includegraphics[width=0.49\textwidth]{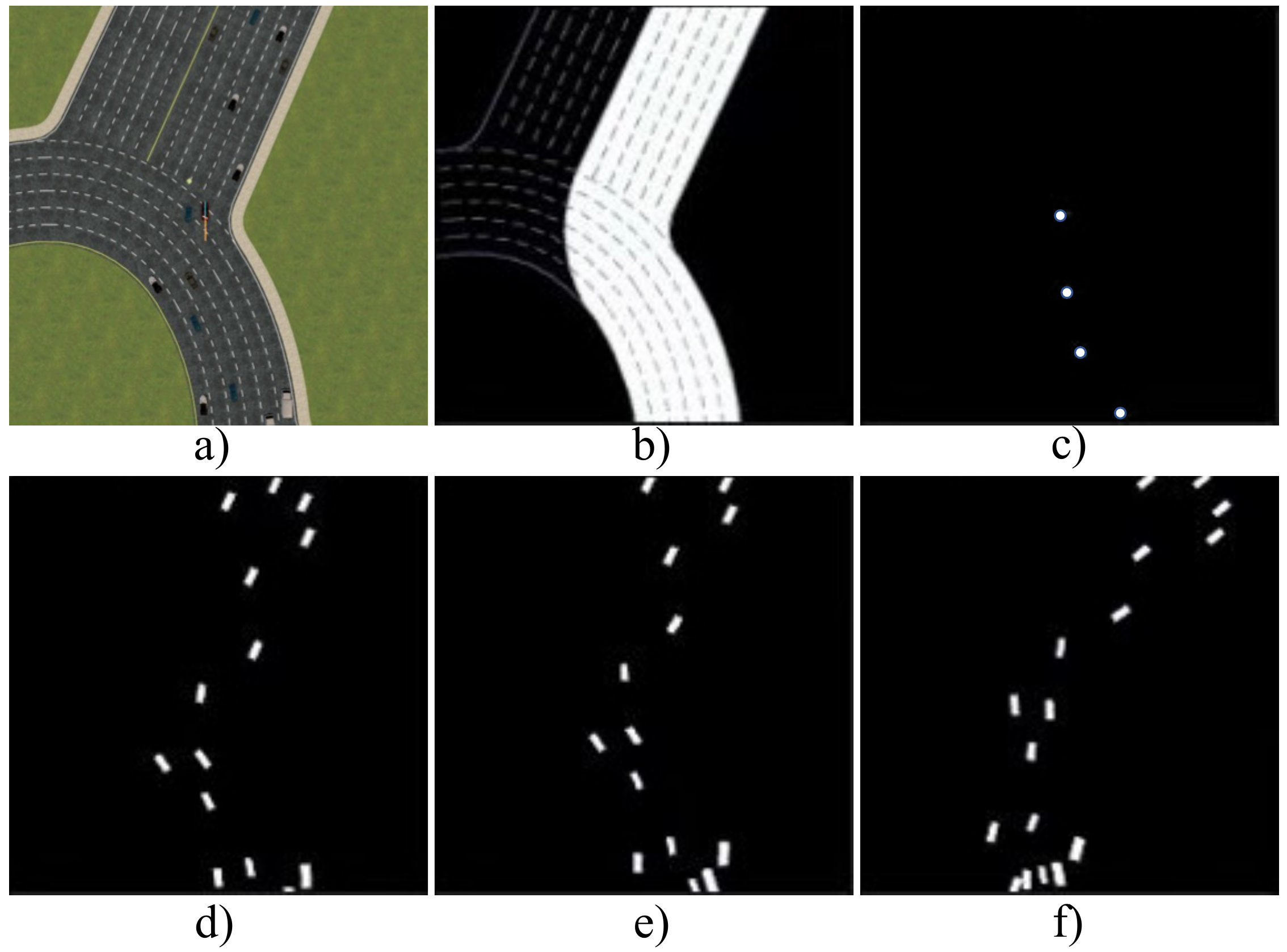}
\caption{Illustration of 5-channel Bird-eye View Observation. Sub-fig(a) is the fragment of a driving scene and (b-f) are 5-channel observation tensors. (b) Road information and target lane. The light part denotes the navigation goal, and the dash line denotes the road shape. (c) Historical trajectory way-points of the ego vehicle. (d-f) Neighboring occupancy grid map at time t, t-1 and t-2, the white rectangle means the neighboring vehicles.}
\label{fig:observation}
\end{figure}

We target the application of autonomous driving and test the generalizability of TaEc-RL across diverse environments with different traffic settings, conducted on the MetaDrive simulator \cite{li2021metadrive}. As shown in Fig.~\ref{fig: case study}, we consider three common-yet-challenging traffic scenarios: highway, roundabout and intersection. 
The ego vehicle needs to drive to destination within the stipulated time, without collision and out of bounds. 
We use a 5-channel bird-eye
view image as observation tensor with a size of 200 × 200 × 5, as illustrated in Figure \ref{fig:observation}.

\subsubsection{Reward and Step Information} At each simulation step, the ego vehicle choose a control input $\mathbf{u}$ as action, and the reward can be represented as:
\begin{equation} \label{eq:reward}
    r_t = c_1 \cdot R_{driving} + c_2 \cdot R_{speed} + c_3 \cdot R_{termination} + c4 \cdot R_{jerk} 
\end{equation}
\begin{enumerate}
\item The driving reward $R_{driving} = d_t - d_{t-1}$, wherein the $d_t$ and $d_{t-1}$ denote the longitudinal coordinates of the ego vehicle in the current lane of two consecutive time steps,å encourages agent to move forward.
\item $R_{speed} = v_t / v_{max}$, encourages the car to move as fast as possible
\item $R_{termination}$ contains a set of sparse rewards. The reward scheme is positive if the car succeed to drive to destination, negative if the car run out of the road or crash other objects, and zero If the game is not terminal.

\item $R_{jerk}$ measures the stability of the car's motion. The more stable of the trajectory, the less penalty will get.
\end{enumerate}
Note that in TaEc RL pipeline, one environment step consists of $T$ simulation steps. Once a skill is chosen, $T$ actions are executed before sampling the next skill, and the reward will be the summary of $T$-step rewards $\widetilde{r} = \sum_{t=1}^{T}r_t$.
\begin{figure*}[t!]
    \centering
    \includegraphics[width=0.9\textwidth]{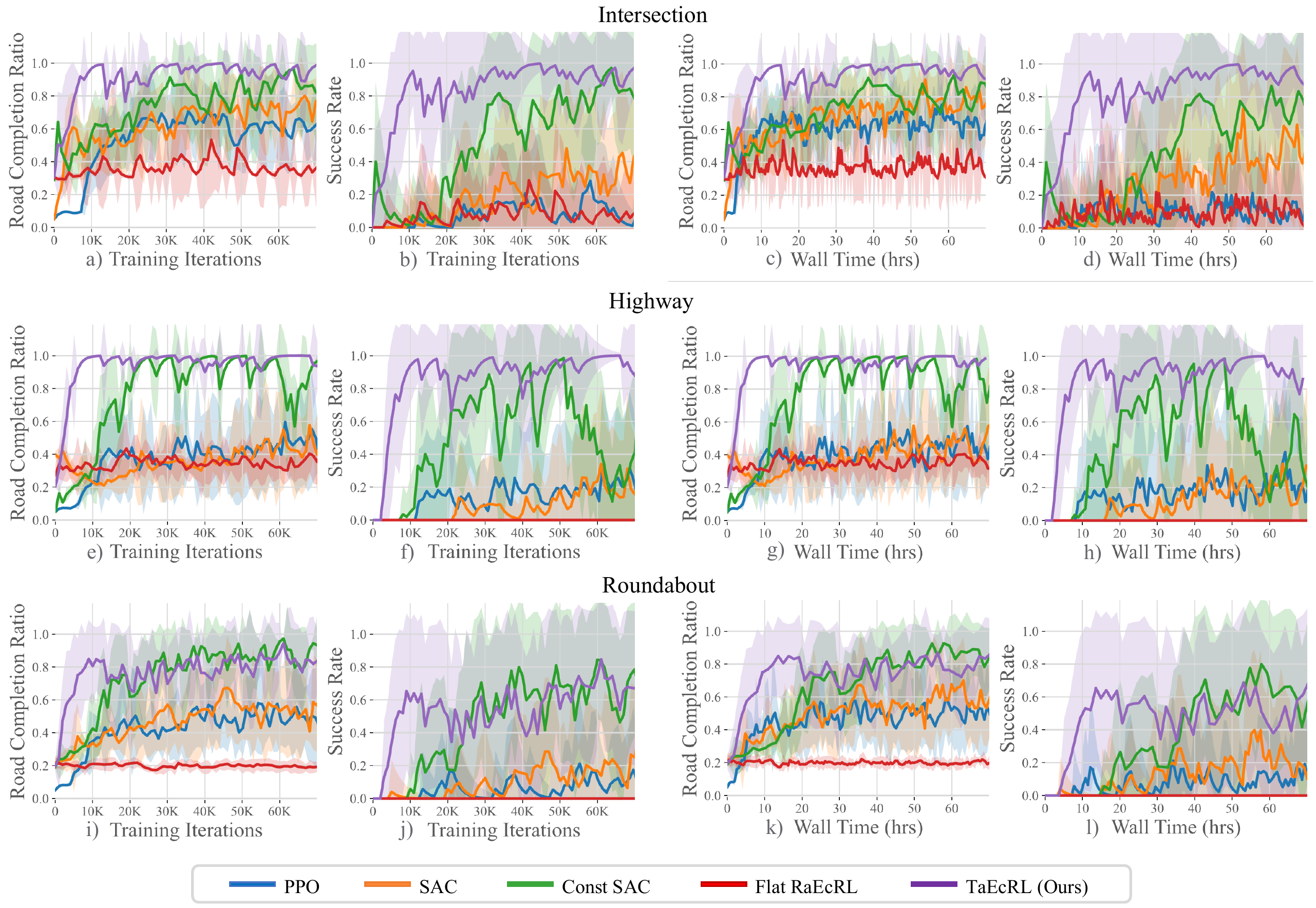}
    \caption{Comparison of our method with other baselines. The driving difficult increases from Intersection, to Highway and Roundabout. SAC and PPO represents the plain RL over raw action spaces. SAC constant temporally extends the same action, and Flat TaEcRL exploits the motion skill of a single step. Our methods exploits both temporal abstraction and motion skill. Compared to plain RL methods, approaches with temporal abstraction (SAC constant) or motion skills (Flat TaEcRL) both show better learning efficiency and task performance as the task become more complex, and TaEc-RL outperformed all other methods. The difference between our TaEc-RL and other methods also becomes more obvious as the task gets more difficult.}
    \label{fig:Training curve}
\end{figure*}

\subsubsection{Baselines} We compare the performance of the proposed TaEc-RL with several baselines:
\begin{itemize}
    \item \textbf{PPO}: Train an agent from scratch by Proximal Policy Optimization (PPO) \cite{DBLP:journals/corr/SchulmanWDRK17}.
    \item \textbf{SAC}: Train an agent from scratch with Soft Actor-Critic (SAC) \cite{haarnoja2018soft}. Along with PPO, they serve as typical RL baseline methods with single-step output for comparison. 
    \item \textbf{Flat TaEcRL}: Train an agent with output of single-step skill. This method tests the effect of temporal abstraction in skills.
    \item \textbf{SAC constant}: Train an agent that output a fixed control signal (once decided) during a period of time $T$ same as the skill horizon of TaRc-RL. This method verifies the necessity of skill diversity and expressiveness with the same decision frequency as TaEc-RL.
\end{itemize}

\subsubsection{Evaluation} 
To evaluate the optimal and stable performance, we used two commonly-used metrics:
\begin{itemize}
\item Success rate: the rate of successfully arriving at the destination within a specified time without collisions and failures;
\item Road completion ratio: the ratio of road length completed against the whole road length
\end{itemize}

Besides, we evaluate the learning efficiency by the following indicators as in \cite{dalal2021accelerating}:
\begin{itemize}
\item Training iterations: the number of backpropagations;
\item Wall time: the total training time (including learning time and interaction time with the environment). 
\end{itemize}

We combine these two sets of indicators in pairs to compare with the baseline.
The skill horizon $T$ for temporally extended methods is set as 10 and the traffic density is set as 0.3.
To ensure the training insistancy, we test all experiments on a single V100 GPU with 16 CPUs and 150G of memory. 

\subsection{Comparisons}
We choose the success rate - training iterations as the main analysis metric-pair as a result of great commonalities in all groups of evaluation indicators. The other three metric-pairs (complete ratio - training iterations, success rate - wall time, complete ratio - wall time) are briefly analyzed. All results are shown in different columns in Fig.\ref{fig:Training curve}.

It can be easily observed that our TaEc-RL algorithm learns a highest success rate fastest among all three scenarios.
It takes about only 10k iterations to reach over $90\%$ success rate in Intersection and Highway scenarios, and $60\%$ success rate in Roundabout.
In addition, our algorithm also reaches the optimum when it converges.
In Intersection and Highway scenarios, our algorithm exceeds all baselines with nearly $100\%$ success rate. And in Roundabout, our algorithm maintains nearly $80\%$ success rate.
Const SAC also achieved relatively good convergence performance while the training iteration required is several times more than ours, and its performance oscillates violently in Intersection and Highway.
TaEcRL Flat performs the worst, especially in Highway and Roundabout scenarios with nearly $0\%$ success rate. This comparison proves the importance of temporal abstraction for RL training in complex scenarios. SAC performed a little better than PPO, but neither achieved $50\%$ success rate.

The trend of results for the metric of road completion ratio is basically the same as that of the success rate. Our algorithm basically converges at 10k iterations, with $100\%$ complete ratio in Intersection and Highway scenarios and $80\%$ complete ratio in Roundabout. Const SAC performs similar as ours but it has slower convergence speed and larger vibration. TaEcRL Flat has similar road completion ratio as SAC and PPO in Highway, but still performs worst in the other two scenarios.

The third and fourth column of Fig.\ref{fig:Training curve} shows the comparison of road completion ratio and success rate with respect to wall time. Most of the results are similar to the previous discussion, except that SAC has a short rise on success rate in Intersection and Roundabout after 50 hrs, but still not comparable to our method.

Through the above analysis, it can be concluded that our algorithm has the fastest convergence in each scenario, and achieves the optimal or near-optimal performance.

\subsection{Visualizations and ablation study}
\begin{figure}[ht]
    \centering
    \includegraphics[width=0.45\textwidth]{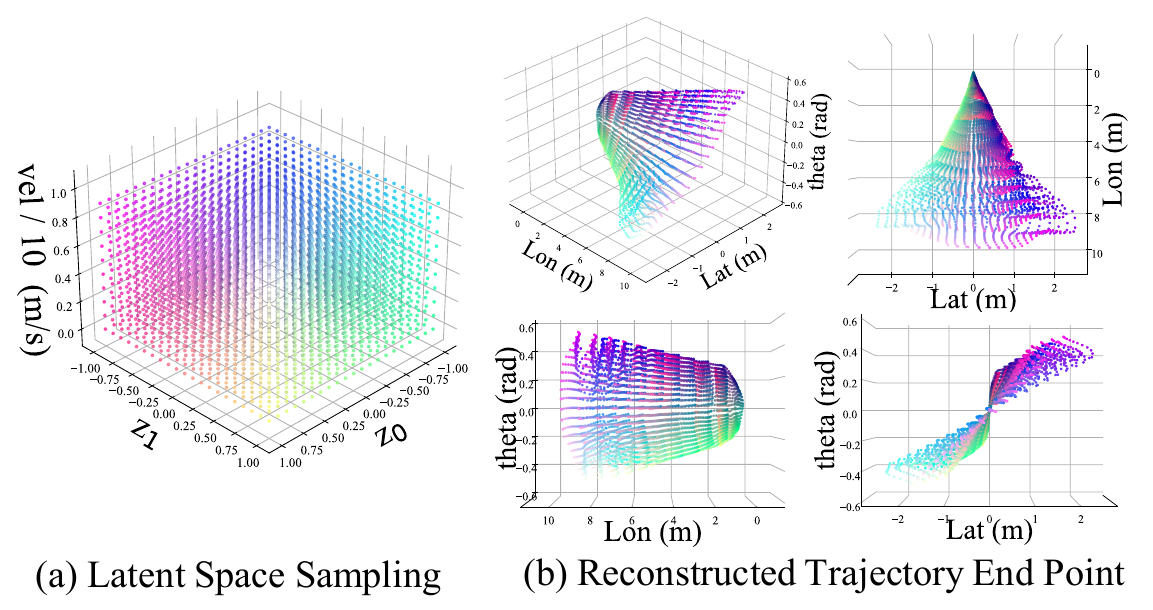}
    \caption{Visualization of latent skill space. Fig (a) shows samples in the latent skills with latent dimension size 2 and horizon length 10, together with the ego vehicle's speed which is also the input of RL policy. Fig (b) visualizes the corresponding reconstructed motion skill end point denoted with the same color in (a).}
    \label{fig:Latent space sampling}
\end{figure}

We deploy visualization of motion skill library by sampling evenly in the latent skill space, and draw the distribution of its correlated trajectory end point. As in Fig.~\ref{fig:Latent space sampling}, diverse motion skills can be reconstructed from the latent skill. Such visualizations also provide interpretability of the latent skill space. The latent skill space is a complete and continuous representation of various motion skills in hyperspace dimensions.

\begin{figure}[ht]
    \centering
    \includegraphics[width=0.45\textwidth]{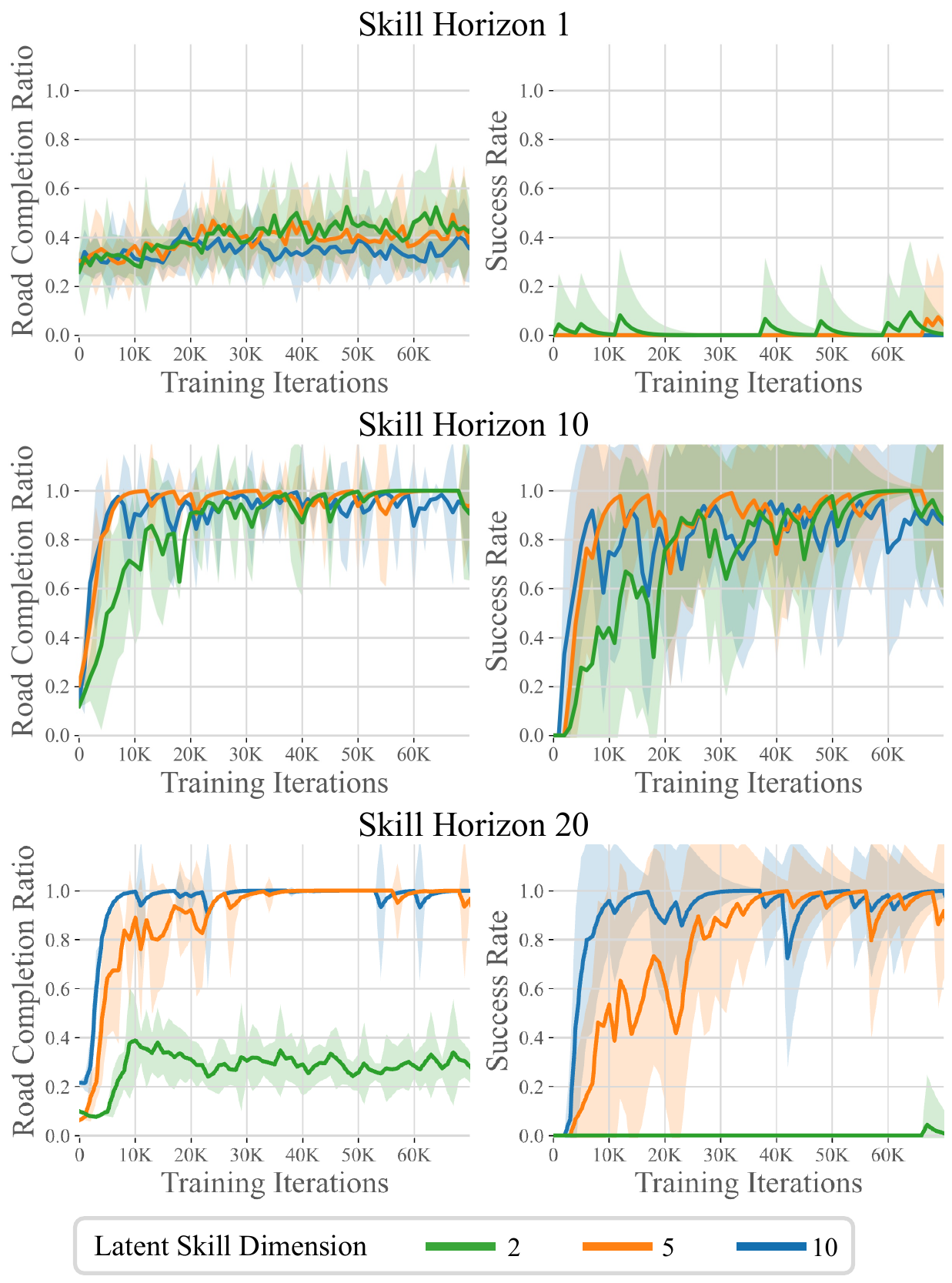}
    \caption{Ablation analysis of skill length and hidden dimension in Highway scenario. Different horizon scenarios (length 1, 10, 20) are listed from top to down. At each horizon scenario, three sets of latent dimension (2, 5, 10) are compared with respect to road completion ratio and success rate.}
    \label{fig:Ablation study}
\end{figure}

In order to get the best skill horizon and latent dimension, we deploy two sets of comparative experiments. Detailly, three sets of horizon with length 1, 10, 20 and three sets of latent dimension 2, 5, 10 are tried and shown in Fig.~\ref{fig:Ablation study}. A short skill horizon (Skill Horizon 1) has low success rate because it loses the temporal coherence of a driving skill. As the horizon becomes longer, the convergence speed of the low skill dimension gets slower, which illustrates that longer horizon trajectories need higher skill dimensions to represent.

Also, long skill horizon fails in small latent skill dimensions, due to the limitation of expressiveness of latent space. Finally, latent skill dimension 5, and skill horizon 10 is chosen to have the fastest convergence with the highest success rate.

\section{CONCLUSIONS}
We present TaEc-RL, an RL method over motion skills to solve diverse and complex driving tasks without demonstration. We design Task-agnostic and Ego-centric motion skill library to cover diverse motion skills. The motion skills are distilled into a latent skill space by a reconstruction process. The RL algorithm is modified to explore in the skill space rather than raw action space. Validations on three challenging dense-traffic driving scenarios demonstrate that our TaEc-RL significantly outperforms its counterpart especially when the driving task become more complex.

\bibliographystyle{IEEEtran}
\bibliography{ref.bib}
\end{document}